\documentclass[letterpaper, 10 pt, conference]{ieeeconf}  % Comment this line out if you need a4paper

\IEEEoverridecommandlockouts

\usepackage{tcolorbox}

\usepackage{graphicx} % for pdf, bitmapped graphics files
\usepackage{amsmath} % assumes amsmath package installed
\usepackage{amssymb}  % assumes amsmath package installed
\usepackage{xspace}
\usepackage{url}
\usepackage{array}
\usepackage{booktabs}

\usepackage{hyperref}
\usepackage{subfigure}
\usepackage{bytefield}
\usepackage{multirow}
\usepackage{tikz}
\usepackage{rotating}
\usetikzlibrary{positioning, shapes, arrows, automata}
\usepackage{glossaries}
\glsdisablehyper % Do not create hyperlinks for gls
\usepackage[normalem]{ulem}
\usepackage{fancyhdr}
\usepackage{marginnote}

\fancypagestyle{withfooter}{
  
  \fancyfoot[C]{\footnotesize Accepted to the IEEE ICRA Workshop on Field Robotics 2025}
}

\newacronym{fm}{FM}{foundation model}
\newacronym{llm}{LLM}{large language model}
\newacronym{vlm}{VLM}{visual language model}
\newacronym{uav}{UAV}{unmanned aerial vehicle}
\newacronym{ugv}{UGV}{unmanned ground vehicle}
\newacronym{vln}{VLN}{visual language navigation}
\newacronym{swap}{SWaP}{size, weight and power}

\hypersetup{
    colorlinks,
    linkcolor={red!50!black},
    citecolor={blue!50!black},
    urlcolor={blue!80!black}
}

\newif\ifdraftcolor
\ifdraftcolor
\newcommand{\fer}[1]{{\color{orange}#1}}

\newcommand{\todo}[1]{{\color{red}#1}}

\newcommand{\deletecamready}[1]{{\color{Red}\sout{#1}}}
\newcommand{\authorlist}{Zachary Ravichandran*,
Fernando Cladera*,
Jason Hughes,
Varun Murali,
\\
M. Ani Hsieh,
George J. Pappas,
Camillo J. Taylor,
and Vijay Kumar}
\else
\newcommand{\fer}[1]{#1}

\newcommand{\todo}[1]{}

\newcommand{\deletecamready}[1]{}
\newcommand{\authorlist}{Zachary Ravichandran*,
Fernando Cladera*,
Jason Hughes,
Varun Murali,
\\
M. Ani Hsieh,
George J. Pappas,
Camillo J. Taylor,
and Vijay Kumar}
\fi

\newcommand{\citationneeded}[1]{$^{[\text{\color{blue}citation needed}]}$~}

\title{\LARGE \bf
Deploying Foundation Model-Enabled \fer{Air and Ground} Robots \\in the Field: Challenges and Opportunities
}

\author{\authorlist%
\thanks{* denote equal contribution.}
\thanks{All authors are with the GRASP Laboratory, University of Pennsylvania.
Corresponding authors: \texttt{{zacravi, fclad} @ seas.upenn.edu}. We gratefully acknowledge the support of
ARL DCIST CRA W911NF-17-2-0181, % DCIST (multirobot)
NIFA grant 2022-67021-36856, %anything to do with agriculture/forestry
the IoT4Ag Engineering Research Center funded by the National Science Foundation (NSF) under NSF Cooperative Agreement Number EEC-1941529, %anything to do with agriculture/forestry
NVIDIA, %
and the NSF Graduation Research Fellowship Program.
}
}

\begin{document}

\maketitle
\thispagestyle{withfooter}
\pagestyle{withfooter}

%%%%%%%%%%%%%%%%%%%%%%%%%%%%%%%%%%%%%%%%%%%%%%%%%%%%%%%%%%%%%%%%%%%%%%%%%%%%%%%%
\begin{abstract}
The integration of \glspl{fm} into robotics has enabled robots to understand natural language and reason about the semantics in their environments.
However, existing \gls{fm}-enabled robots primary operate in closed-world settings, where the robot is given a full prior map or has a full view of its workspace.
\fer{This paper addresses the deployment of \gls{fm}-enabled robots in the field,} where missions often require a robot to operate in large-scale and unstructured environments.
\fer{To effectively accomplish these missions}, robots must actively explore their environments, navigate obstacle-cluttered terrain, handle
\fer{unexpected sensor inputs, and operate with compute constraints.}
We discuss recent deployments of SPINE, our LLM-enabled autonomy framework, in field robotic settings.
To the best of our knowledge, we present the \emph{first demonstration of large-scale LLM-enabled robot planning in unstructured environments} with several kilometers of missions.
SPINE is agnostic to a particular LLM, which allows us to distill small language models capable of running onboard \gls{swap} limited platforms.
Via preliminary model distillation work, we then present the
\emph{first language-driven UAV planner using on-device language models}.
We conclude our paper by proposing several promising directions for future research.

\end{abstract}

%%%%%%%%%%%%%%%%%%%%%%%%%%%%%%%%%%%%%%%%%%%%%%%%%%%%%%%%%%%%%%%%%%%%%%%%%%%%%%%%

\glsresetall

\section{Introduction}
The integration of \glspl{fm} into robotics --- namely, \glspl{llm} and \glspl{vlm} --- has provided robots with impressive contextual-reasoning capabilities.
\glspl{fm} allow robots to reason over both mission specifications expressed in natural language and complex semantic associations in their environments.
Roboticists have leveraged these capabilities to design
\gls{fm}-enabled planners for mobile manipulation~\cite{rana2023sayplan}, autonomous driving~\cite{li2024llada}, and service robots~\cite{llm_service_robot}, among other applications.
These planners typically leverage the contextual knowledge of FMs by translating a natural-language specification into plans parameterized by APIs~\cite{ravichandran_spine}, code~\cite{codeaspolicies2022}, or a formal language~\cite{liu23lang2ltl}, which can be executed by the robot.

\begin{figure}[ht!]
    \centering
    \includegraphics[width=1\linewidth]{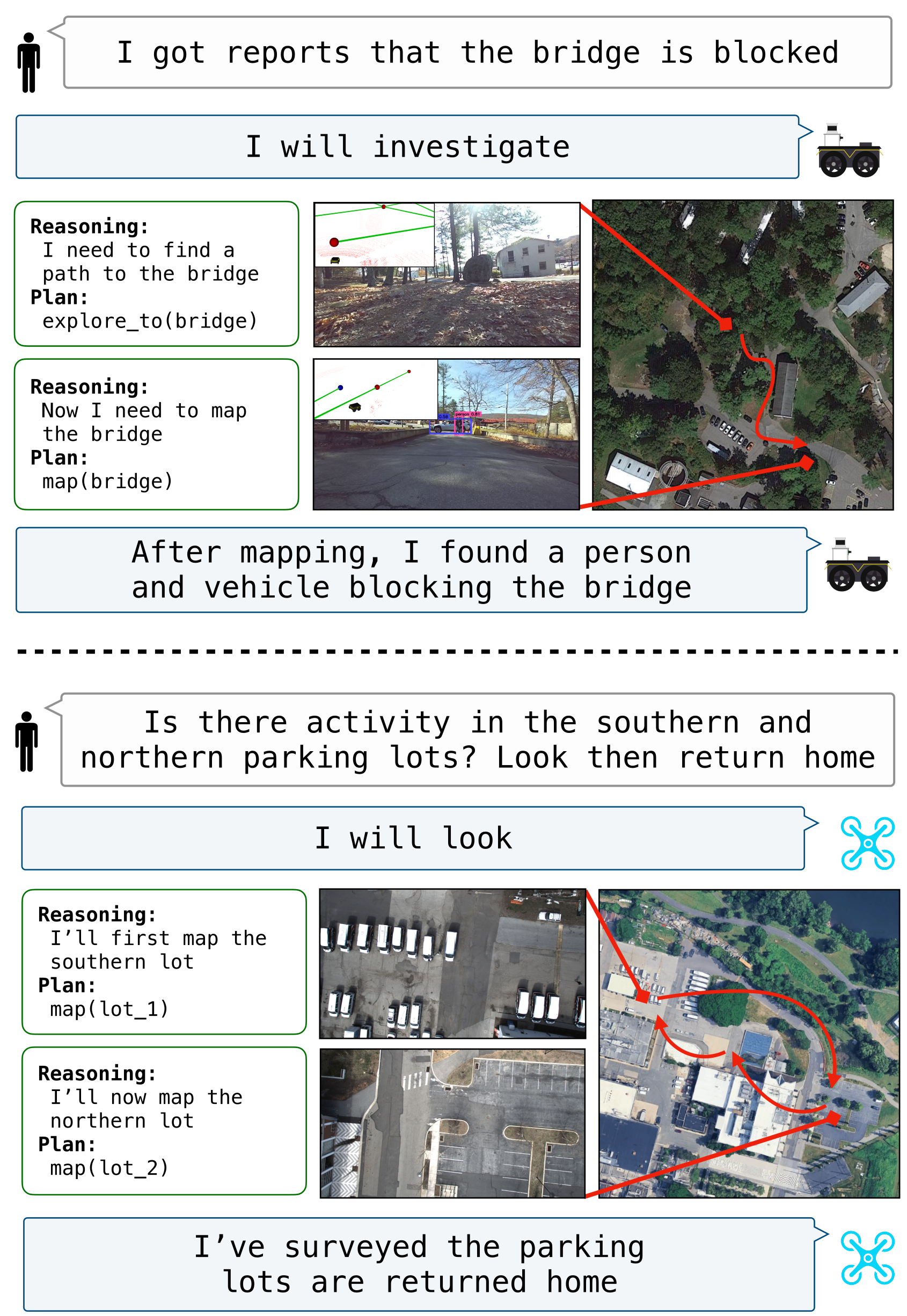}
    \caption{Example missions. \uline{Top}: The UGV is given a language-specified mission in a partially-known environment, and it performs several steps of reasoning, planning, and active mapping in order to fulfill the user's request. \uline{Bottom}: Deployment of a distilled language model running \textit{onboard} a UAV for planning.}
    \label{fig:fig1}
    \vspace{-.3cm}
\end{figure}

Current FM-enabled robot planners assume the world is relatively structured and known \textit{a priori}.
Planners take a near-complete map at runtime and do not need to consider information acquired  online.
As such, online planning via active perception is relatively unexplored in FM-enabled robotic contexts.
These assumptions are reasonable in many problem domains, such as indoor service robots.
However, field robots operate in uncertain and unstructured environments, which limits the utility of current FM-enabled robots.

This paper discusses our deployment of \gls{fm}-enabled robotics in the field over the past year.
Our deployments focus on missions with \textit{incomplete specifications} in \textit{natural language} -- where the robot must infer subtasks from high-level specifications -- in unknown or partially known environments.
A key differentiating factor of our autonomy framework is that it embeds FMs in a \textit{closed-loop} system, which enables real-time plan verification and online responses to map updates.
Our experiments consider two FM-enabled planning paradigms.
The majority of our experiments leverage frontier server-based LLMs, namely GPT-4o.
This approach enables planning over complex mission specifications at scales of up to 1 kilometer but requires significant communication infrastructure.
We then present preliminary work that mitigates the communication requirements of server-based LLMs via \textit{distilling} small language models (SLMs) capable of running onboard robot compute.
We use this to present the first (to the best of our knowledge) demonstration of language-based UAV planning with fully onboard computation.
Finally, we discuss lessons learned during our deployments and propose open challenges in \gls{fm}-enabled field robotics.
In summary, our contributions are:
\begin{itemize}
    \item We introduce the \textbf{first demonstration of a large-scale LLM-enabled robot} planning in \textbf{unstructured environments}, achieving kilometer-scale missions.
    \item We present the \textbf{first-time deployment of distilled light-weight models on aerial robots}.
    \item We present lessons learned on our field experiments, as well as open challenges in deploying LLM-enable robots in the field.
\end{itemize}

\subsection{Related work}

\noindent \textbf{FM-enabled robots}
Researchers have applied FM-enabled planners to domains including mobile manipulation~\cite{momallm24, rana2023sayplan}, service robots~\cite{llm_service_robot}, autonomous driving~\cite{li2024llada}, and navigation~\cite{xie2023reasoning, RoboHop, pmlr-v229-shah23c, pmlr-v205-shah23b}.
FM-enabled planners typically configure pre-trained model via prompts which include a problem description, robot interface, and specification.
These approaches enable complex reasoning, but they typically provide the planner with a \emph{near complete} map.
And while some FM-enabled planners incorporate feedback from perception systems~\cite{momallm24, huang2022inner}, perception is limited to 2D representations (\textit{e.g.}, object detection) or designed for small room-centric environments.
A second limitation of these works is that they typically operate in highly structured environments, such as homes or offices.
A few works consider large-scale outdoor navigation~\cite{pmlr-v205-shah23b, Shah-RSS-22}, they still require a prior maps to represent goal locations or navigation graph.
\medskip

\medskip

\noindent \textbf{FM-enabled Aerial robots. } Larger scale demonstrations of foundation models on \glspl{uav} have focused on simulations~\cite{cai2024neusis} where the compute requirements are relaxed, and the nature of closed-loop interaction with a noisy autonomy system is simplified.
\cite{liu_2023_AerialVLN} generates an outdoor \gls{vln} dataset similar to the indoor \cite{he2021landmark} dataset that employs simulation with human-annotated navigation trajectories for conversationally specified tasks.
\cite{gao2024aerial} generates a physically grounded semantic-topological map from aerial imagery to incrementally construct a map and query an
\gls{llm}-based planner to generate the next discrete action --- turn left or right, go forward, or stop.
While they deploy perceptual components on the \gls{uav}, they communicate with an offboard GPT-4o model.
NEUSIS develops a neuro-symbolic framework that decomposes the task into physically grounded perceptual tasks like detecting objects and mapping them in 3D space, similar to~\cite{ravichandran_spine}.
However, they primarily focus on simulation and tasks that are well-specified in both the primary target of interest and the area of interest.

\section{Autonomy  Framework}

\begin{figure}
    \centering
    \includegraphics[width=1\linewidth]{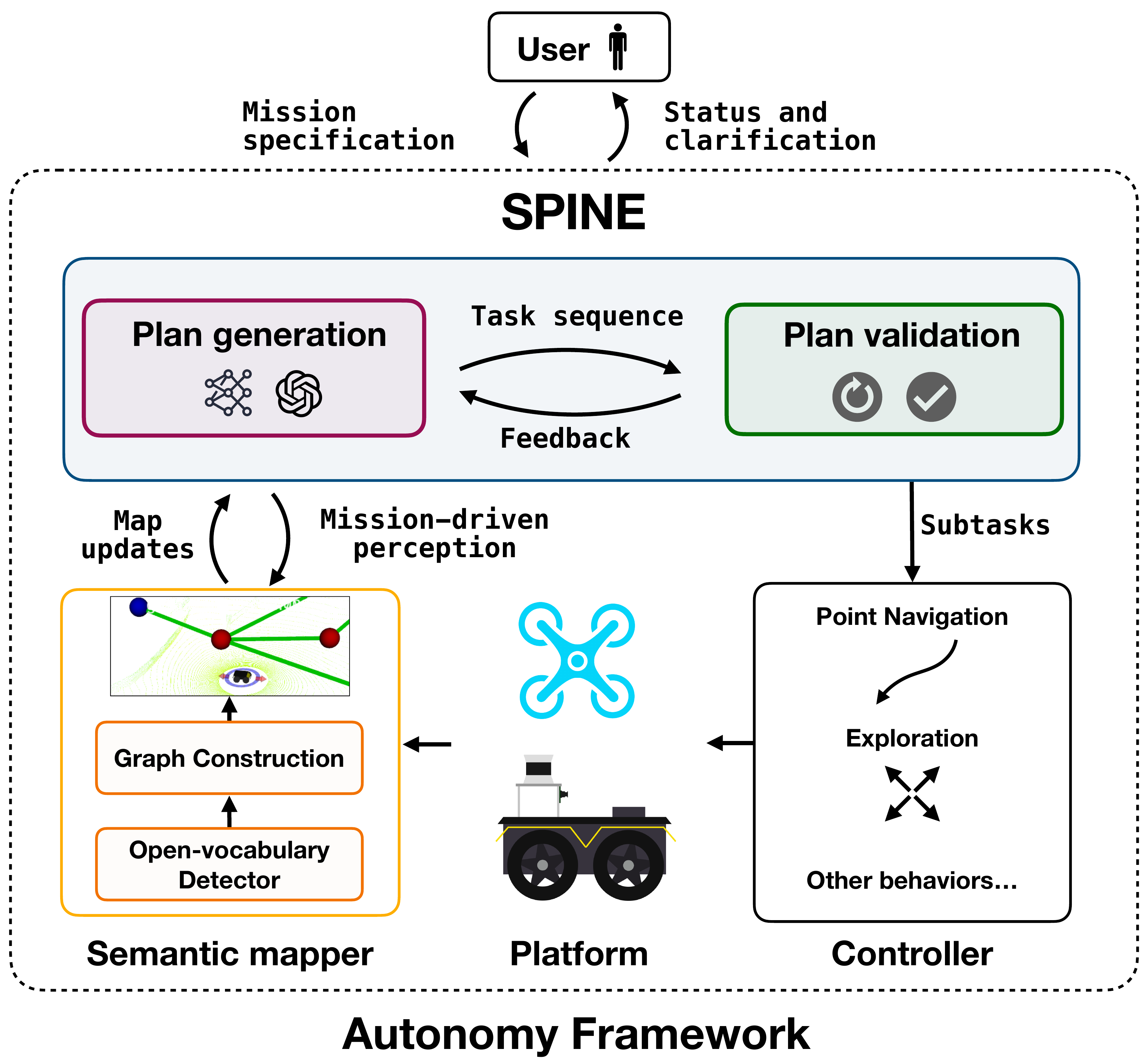}
    \caption{UGV/UAV autonomy overview. Given a language-specified mission, SPINE uses an LLM to generate an appropriate task sequence, which is validated online for physical safety and realizability.
    These subtasks are executed by a controller capable of performing atomic behaviors such as point navigation or exploration. The planner can also configures a mapping framework, from which it receives real-time updates. Adapted from \cite{ravichandran_spine}. }
    \label{fig:spine_arch}
    \vspace{-.3cm}
\end{figure}

We consider a robot operating in an environment that is \emph{unstructured} and either \emph{unknown} or \emph{partially-known}; the robot is capable of performing behaviors such as point navigation and exploration.
The robot is  equipped with a  mapper that continuously builds a semantic map given robot sensory data, and it may also be given a partial prior \fer{semantic map (extracted from satellite images, for instance)}.
The user provides the robot with an \emph{incomplete} mission specifications in natural language, which means the specification implies subtasks that the robot must infer.
Our \gls{llm}-enabled planner, SPINE, interacts with the robot's control and mapping modules in a \emph{closed-loop manner} to execute the user's mission~\cite{ravichandran_spine}.
Our experiments consider  Clearpath Jackal \glspl{ugv} and Falcon 4 \glspl{uav}. Our framework is compatible with any platform that meets SPINE's requirements.
Our autonomy framework is illustrated in Figure~\ref{fig:spine_arch}, and we subsequently  provide an overview of SPINE, each platforms autonomy stack, and the corresponding hardware configuration.

\subsection{SPINE}

SPINE is comprised of two major modules --- \emph{plan generation} and \emph{plan validation} --- we briefly discuss each below, and we refer the reader to \cite{ravichandran_spine} for further details.

\medskip

\noindent \textbf{Plan Generation.} The plan generator takes as input the user's mission specification  and the current semantic map, and it produces task sequence that will fulfill that mission.
Given the reasoning capabilities of frontier LLMs, we use them to instantiate this module.
We configure the LLM to plan over the robot's API as defined by the control module; our implementation uses behaviors for navigation, active mapping, and user interaction, although these may comprise any valid robot behavior.
The plan generator then composes a task sequence using this API via \textit{chain-of-thought reasoning}, which requires the LLM to reason about its plan along with providing the task sequence.
At each planning iteration, the plan generator is provided map updates via a textual interface, which it may use to refine its subsequent plan.

\medskip

\noindent \textbf{Plan Validation.}
Given that LLMs may hallucinate, especially in unstructured settings, SPINE validates LLM-proposed plan before sending them to the control module.
Specifically, each behavior in the robot's control library defines constraints that must be respected when invoking that behavior, and these constraints are of type \emph{syntax}, \emph{reachability}, \emph{explorable}.
Syntax constraints simply require that a function be invoked with the correct arguments.
Reachability and explorable constraints effectively ground LLM-generated plans in the physical world; navigation goals by reachable in the current graph and exploration goals must command an obstacle-free path.
Should the plan generator provide an erroneous task sequence, the plan validator will give feedback in natural language that references the specific violation, which the plan generator may use to fix its plan.

\begin{figure}
    \centering
    \includegraphics[width=0.95\linewidth]{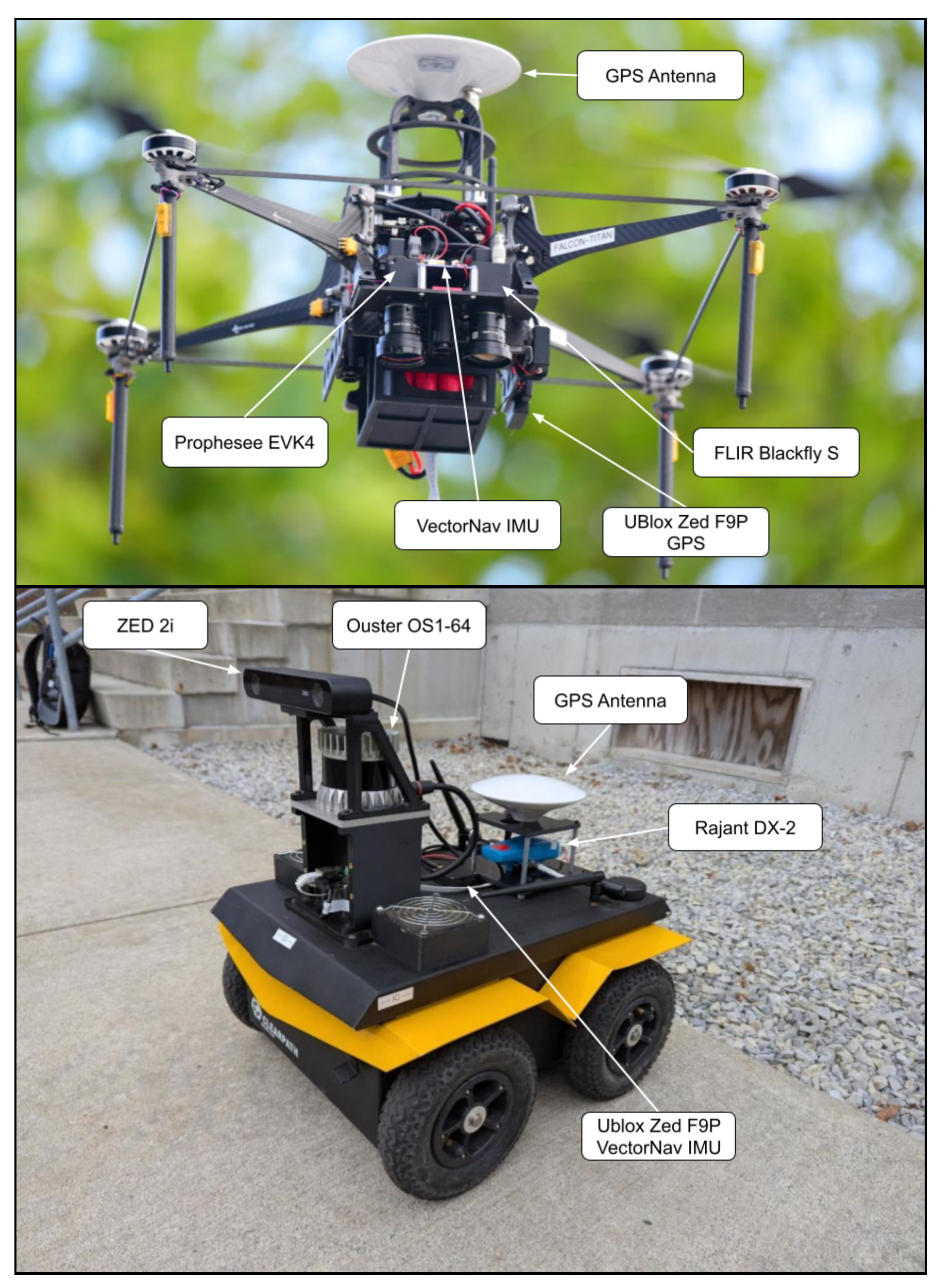}
    \caption{Overview of the robotic platforms and sensors used in our experiments. \uline{Top}: high-altitude Falcon 4 UAV. \uline{Bottom}: Clearpath Jackal. Figure from~\cite{cladera2025tfr}.}
    \label{fig:platforms}
    \vspace{-.3cm}
\end{figure}

\subsection{Autonomy Stack}
The autonomy stack takes SPINE's plan and execute it on the platform.
Additionally, the stack may provide feedback to SPINE to close the loop on the mission specification. Figure~\ref{fig:platforms} shows the platforms used for our experiment.

\noindent \textbf{Ground Autonomy Stack}
We consider Clearpath Jackals fitted with a ZED 2i depth camera, an Ouster OS1 64 LiDAR, a ZED-F9P GPS module, and a VectorNav VN100-T IMU. The onboard computer features an  AMD Ryzen 3600 CPU (32 GB RAM), and an Nvidia RTX 4000 SFF Ada Generation GPU (20 GB VRAM).
The ground autonomy stack uses Faster-LIO for LiDAR Odometry~\cite{bai2022faster} and GroundGrid~\cite{steinke2024groundgrid} to estimate free-space and plan trajectories.
The behaviors used by the semantic planner include \texttt{map\_region}, \texttt{explore\_region}, \texttt{extend\_map}, \texttt{goto}, and \texttt{inspect}.
We refer the readers to~\cite{ravichandran_spine} for more details about the ground autonomy stack.

\medskip

\noindent \textbf{Aerial Autonomy Stack}
The aerial platform is the high-altitude Falcon 4 \gls{uav}~\cite{cladera2024evmapper}, using an ARKV6X flight controller running PX4.
The platform is fitted with a global-shutter RGB camera, a VectorNav VN100-T IMU, and a ZED-F9P GPS. All computation is carried out onboard an Nvidia Jetson Orin NX.
While the platform features an event camera, we do not use it for these experiments.
The aerial autonomy stack is based on \texttt{air\_router}, first introduced in~\cite{cladera2024enabling}. In this work, we only task the UAV with navigating to coordinates, and thus the behaviors available are \texttt{inspect}, \texttt{map\_region}, \texttt{explore\_region}, and \texttt{goto}. The autonomy stack selects a waypoint from a list of pre-defined locations closer to the requested location and flies the \gls{uav} to that waypoint.

\subsection{Communication Infrastructure}
\label{subsec:comms}
One of the challenges of deploying server-based \glspl{fm} is that the robot must have continual access to the model's server.
We leverage mesh networking to ensure continuous communications, provided by Rajant breadcrumb DX-2 and Cardinal radios.
Upstream internet is provided by a 5G cellular link or satellite internet in isolated experiment sites.
To extend the operation range of the robots, we place \emph{dummy} breadcrumb nodes to increase the mesh range.
Nonetheless, the limited operational range motivates the deployment of models on the edge, as described in \S\ref{sec:on_the_edge}.

\section{Deploying Foundation models in the Field}
\label{sec:fm-on-the-field}

\begin{figure}
    \centering
    \includegraphics[width=1\linewidth]{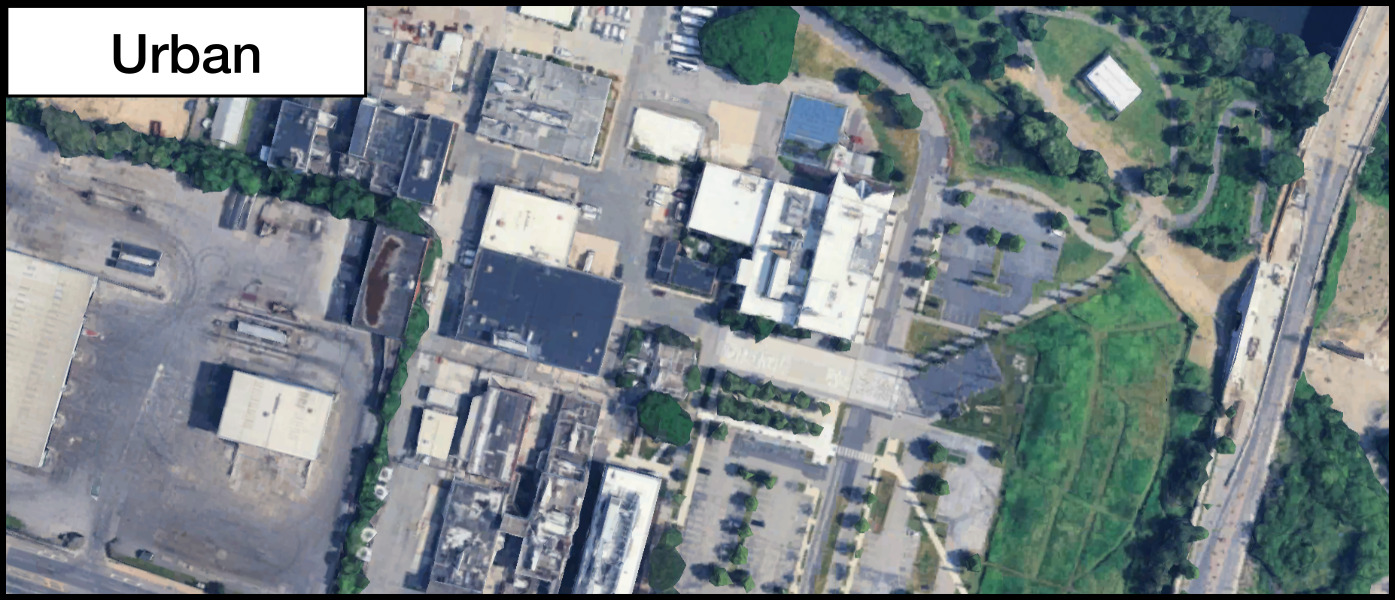}

    \vspace{.1em}

    \includegraphics[width=1\linewidth]{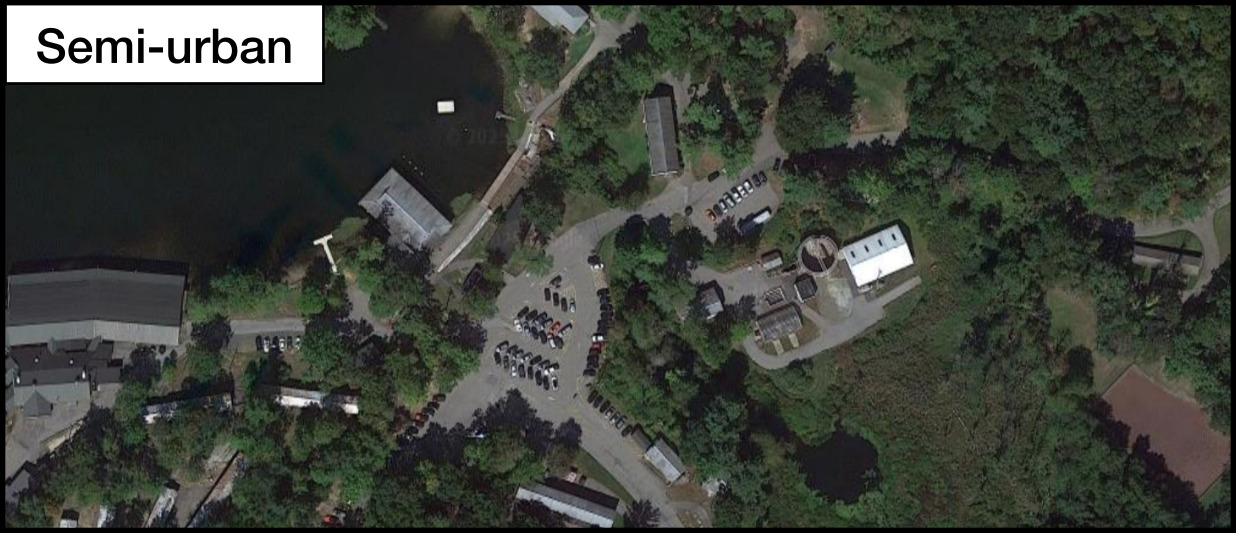}

    \vspace{.1em}

    \includegraphics[width=1\linewidth]{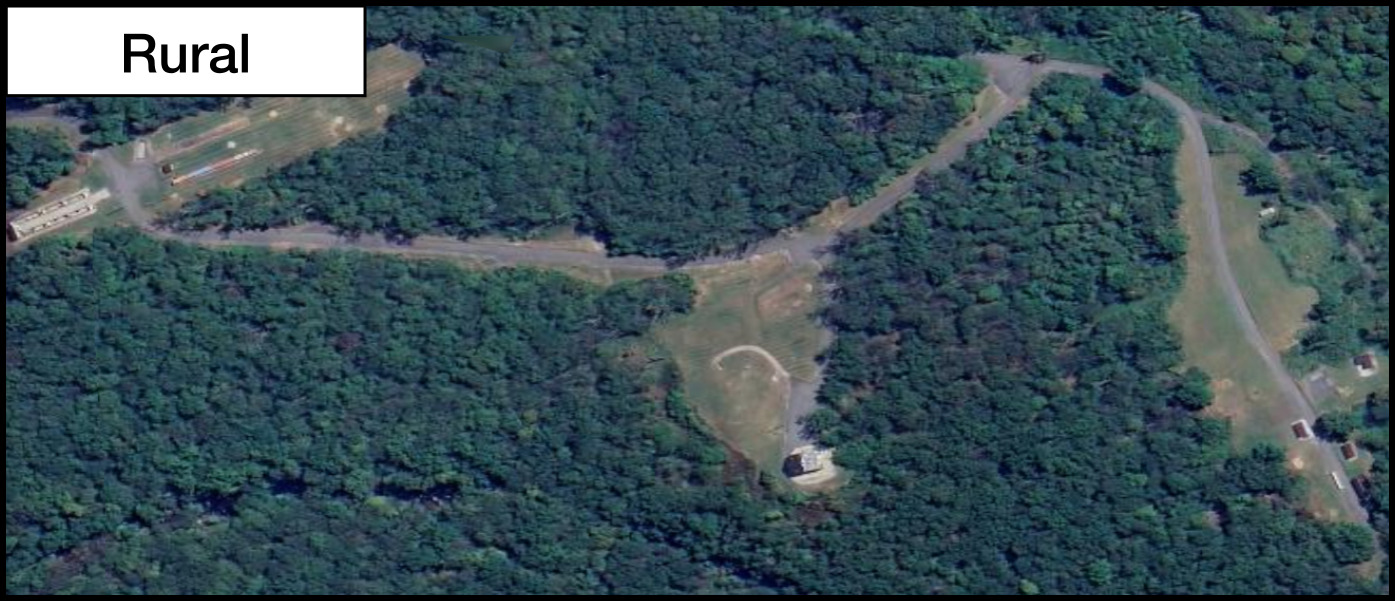}
    \caption{We consider three experimental environments: \uline{Top}: an urban office park. \uline{Middle}: a semi-urban environment. \uline{bottom}: a rural environment (figure adapted from \cite{cladera2025tfr}).}
    \label{fig:envs}
    \vspace{-.3cm}
\end{figure}

We have tested our SPINE-based autonomy architecture in field sites that range from urban office parks to rural environments, as shown in Fig.~\ref{fig:envs}.
The majority of our experiments use the Clearpath Jackal, which we summarize in \S\ref{sec:ground} and refer the reader to \cite{ravichandran_spine, cladera2025tfr} for further detail.
Importantly, these experiments use GPT-4o, which requires internet infrastructure.
In preliminary experiments, we overcome this requirement by distilling an SLM for robot planning, which can run onboard a UAV in real time, discussed in \S\ref{sec:on_the_edge}.

\subsection{Deploying LLM-enabled UGVs}
\label{sec:ground}

\begin{figure}
    \centering
    \includegraphics[width=1\linewidth]{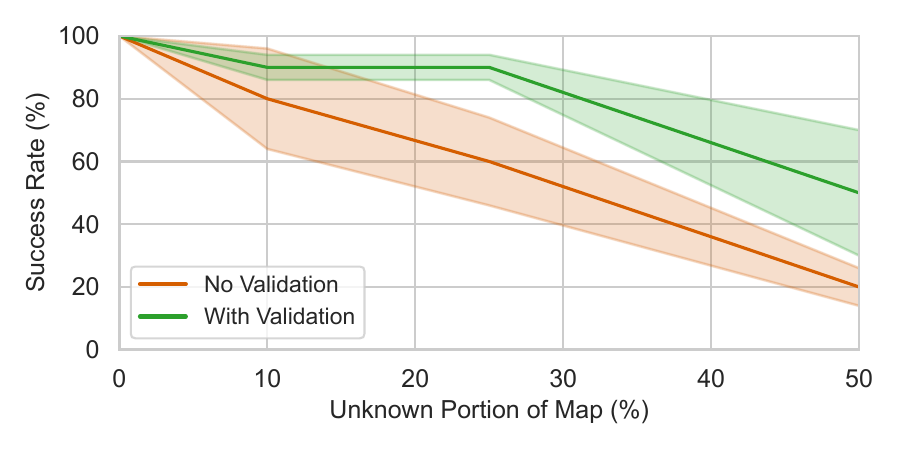}
    \caption{Importance of validation. Even minimal feedback significantly improved performance of LLMs. Figure from \cite{ravichandran_spine}).}
    \label{fig:validation}
    \vspace{-.3cm}
\end{figure}

We deploy our UGV-based autonomy in \emph{urban}, \emph{semi-rural}, and \emph{rural} environments, in which we evaluate SPINE's ability to execute incompletely-specified missions in partially-known environments; these missions require 2-8 steps of nontrivial reasoning while requiring the robot to traverse 100m to 1km.
Tab.~\ref{tab:ugv_outcomes} provides a summary of results over specifications (Sx).
SPINE completed twelve out of the fourteen missions, with failure resulting from communication loss and odometry drift.
Obstacle detection failure occurred during S7 and S8. However, SPINE was able to complete the mission after brief manual takeover.
We highlight the importance of online validation in Fig.~\ref{fig:validation}, without which the mission success rate significantly drops as the environment becomes increasingly unknown.

\begin{table}[b]
    \begin{tabular}{cccc} \toprule
         Specification   & Outcome & Avg. Distance (m) & Failure modes\\ \toprule
         S1 & 1/3 & 1200 & Odometry, Comm. \\
         S2 & 2/2 & 231   & N/A \\
         S3 & 4/4 & 265 &  N/A \\
         S4 & 1/1 & 132   &  N/A \\
         S5 & 1/1 & 450  &   N/A \\
         S6 & 1/1 &  303   &  N/A \\
         S7 & 1/1 &   164  & Obst. det. \\
         S8 & 1/1 &  219  &  Obst. det. \\
        \toprule
    \end{tabular}
    \caption{UGV autonomy outcomes (from \cite{cladera2025tfr} and \cite{ravichandran_spine}).}
    \label{tab:ugv_outcomes}
\end{table}

\subsection{Onboard FM-Enabled Planning: Initial Results \fer{on UAVs}}
\label{sec:on_the_edge}
Using server-side LLMs requires a robust communication infrastructure, which may be unavailable in the field.
LLM distillation --- which trains a smaller model to mimic a larger LLM --- is a promising solution.
Retaining the generalized reasoning capability of the LLM is a challenge of this solution.
Another challenge is data collection --- one of the significant advantages of LLMs is that they require minimal tuning, and an approach that requires extensive data collection would mitigate these advantages.

\medskip

\noindent\textbf{Distilling on-device LMs for robot planning }
We construct a preliminary framework for distilling small language models (SLMs) capable of running onboard a robot (\textit{e.g.}, using under 5 GB, for example).
This framework is comprised of an expert planner, a data collector, and a distillation method.
The data collector creates a dataset from the expert planner, whose behavior the SLM should mimic, by providing the expert with a \verb|specification| and \verb|semantic graph|. Given these input sources, the expert provides a plan comprising \verb|observation| and \verb|action| tuples.

Our preliminary experiments use SPINE equipped with GPT-4o as the expert planner.
We implement the data collector with a secondary instantiation of GPT-4o, which is tasked with generating mission specifications, and we use semantic graphs from previous missions.
The distillation method uses LoRA~\cite{hu2021lora} to finetune Llama-3.2 3B~\cite{touvron2024llama3}.

\medskip

\begin{figure}[t!]
    \centering
    \includegraphics[width=1\linewidth]{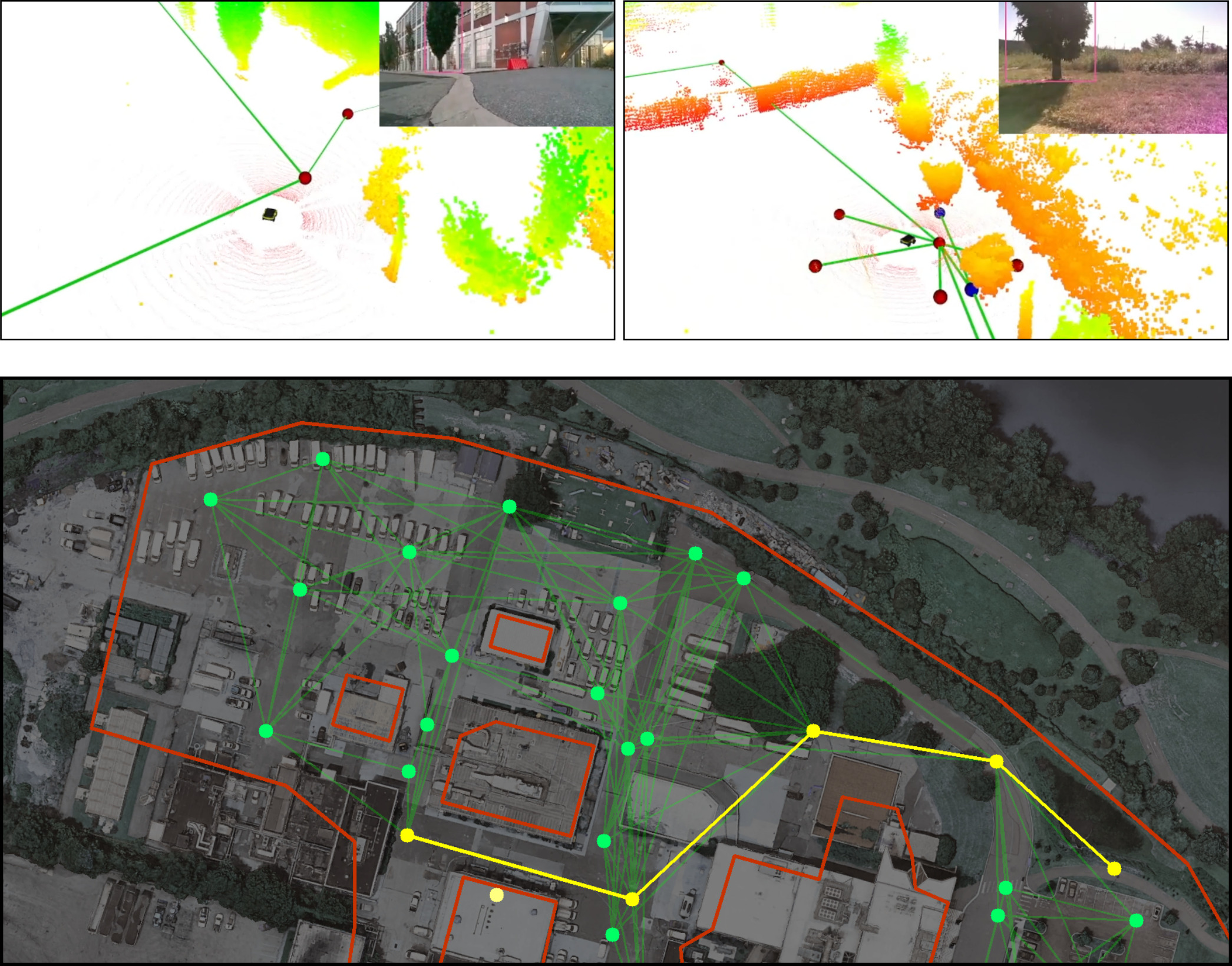}
    \caption{Semantic maps used by our autonomy framework. \uline{Top}: UGV semantic maps. \uline{Bottom}: UAV map (green), geofence (red) and path (yellow). Semantic maps comprise region and object nodes, and provide a common interface between the UAV and UGV.}
    \label{fig:semantic-maps}
\end{figure}

\noindent\textbf{Results.}
We demonstrate the LLM-enabled UAV in the the urban environment shown in Fig.~\ref{fig:envs}.
We provide SPINE with a semantic map extended from previous mission data that contains parking lots and parks.
We then task the UAV with the following language-specified missions:
\begin{enumerate}
    \item [1.] Is there activity in the southern parking lot?
    \item [2.] What is going on in the park?
    \item [3.] I heard of construction in the northern parking lot. Check.
    \item [4.] Return to home
\end{enumerate}
SPINE is able to correctly fulfill three out of the four missions on the first try.
During the third mission, SPINE confused the southern parking lot with the northern one. However, this would easily be corrected by this user with a follow-up command.

\medskip

\begin{table}[h!]
    \vspace{1mm}
    \centering
    \begin{tabular}{cc}
        \toprule
        Model & Success Rate \\
        \midrule
        GPT-4o  &  100 \%\\
        Distilled Llama 3.2 3B & 72.7 \% \\
        Off-the-shelf Llama 3.2 3B & 9.2 \% \\
        \bottomrule
    \end{tabular}
    \caption{Planning ability comparison.}
    \label{tab:llm_resuls}
    \vspace{-0.3cm}

\end{table}

\begin{figure}[t!]
    \centering
    \includegraphics[width=0.95\linewidth]{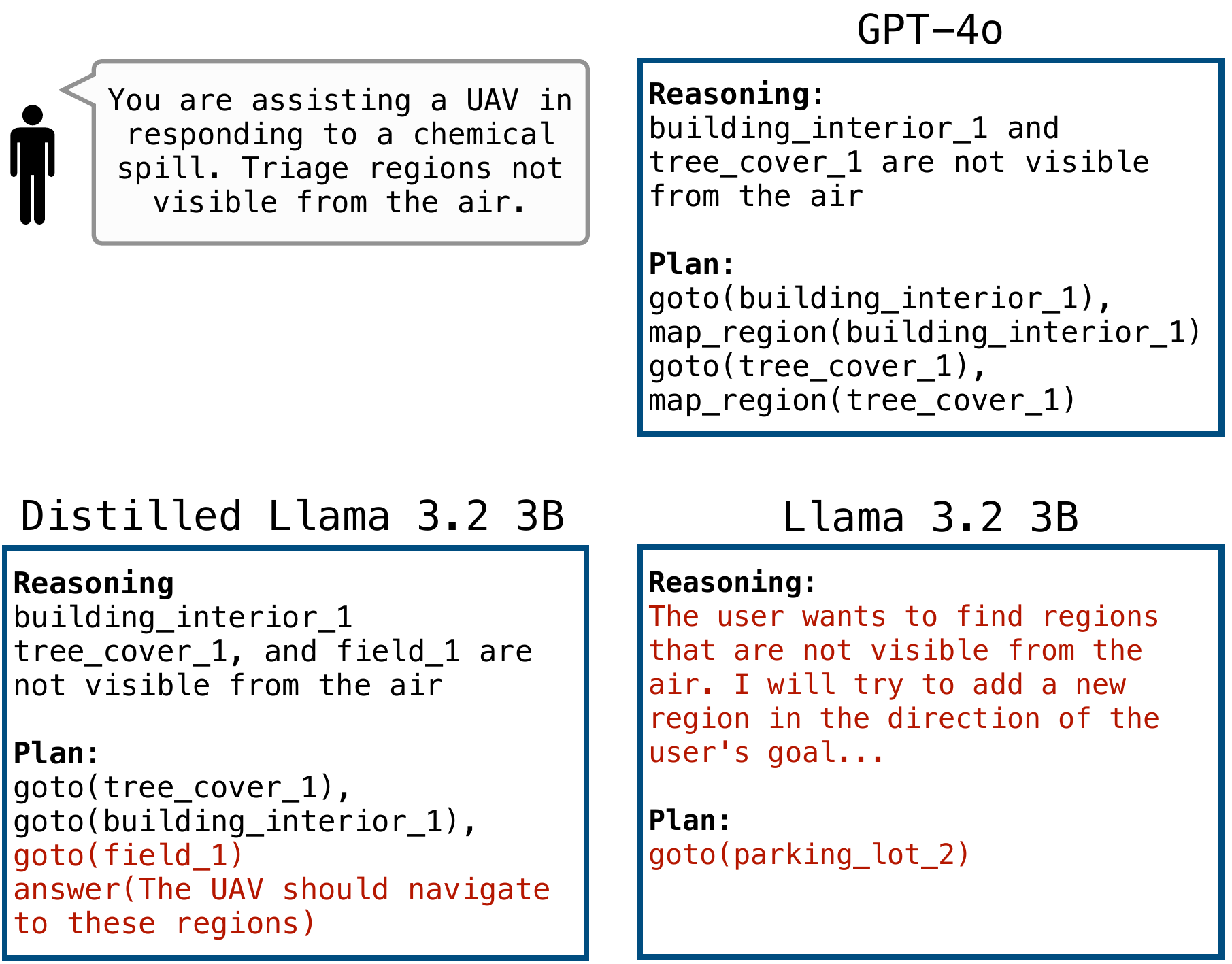}
    \caption{Planning capabilities across language models. each model must plan given a graph with semantics including buildings, fields, trees, and parking lots. \uline{Right}: GPT-4o successfully plans. \uline{Middle}: The distilled Llama 3.1 8B model is nearly correct.
    \uline{Right}: Without distillation, Llama 3.1 8B cannot generate a reasonable plan. Mistakes are in red.}
    \label{fig:large-small-comparison}
    \vspace{-.3cm}
\end{figure}

\noindent\textbf{Failure cases}.
In order to assess the value of distillation,
we compare three models ---  GPT-4o, distilled Llama 3.2 3B, and off-the-shelf Llama 3.2 3B ---  in their ability to correctly produce a plan given eleven different specifications with semantic graphs from previous missions.
To ensure a fair comparison, we provide the off-the-shelf Llama 3.2 3B with the same system prompt and in-context examples as GPT-4o.
A known restriction to our preliminary distillation implementation is its inability to generate long-horizon planning data.
We, therefore, limit evaluation to short-horizon planning.
Tab.~\ref{tab:llm_resuls} reports the results.
Although there is a sizable gap between the distilled LLM and GPT-4o (72.7\% compared to 100\%), distillation still leads to a significant performance gain.
Qualitatively, distillation enabled the model to perform simple reasoning missions and respect the planning API.
However, the distilled model still struggled with multi-iteration planning, often unable to appropriately respond to map updates.
The distilled model also struggled with more complex reasoning missions (see Fig.~\ref{fig:large-small-comparison}).
Nonetheless, these results suggest promise in our distillation approach.

\subsection{Towards Air-Ground Teaming} \label{sec:air-ground-teaming}

Air and ground robots provide unique, separate advantages to a robot team~\cite{miller2022stronger}. A UAV can quickly scan from a high altitude to get a macroscopic overview of the area, while a UGV can get a more detailed view of an area.

In~\cite{cladera2025tfr}, we develop an air-ground teaming framework leveraging foundation vision and language models to complete missions with natural-language specifications.
The core representation of this framework is a hierarchical semantic graph, which is used as a \emph{lingua franca} between the UAVs and UGVs.
%, that allows for robust, task agnostic air-ground team.
As in SPINE~\cite{ravichandran_spine}, a user provides a mission specification in natural language, and the LLM-enabled planner infers  semantic classes required for the mission.
The UAV build a mission-relevant semantic graph using the following procedure.
The UAV uses Grounded SAM2 \cite{ren2024grounded} to detect and localize the inferred classes, along with classes for traversability analysis, such as \texttt{road}.
The UAV then uses these detections to construct a hierarchical graph comprising object and traversability nodes, where edges are defined between traversability nodes and from traversability to  object nodes.
During construction, the graph is periodically sent to the UGV for mission planning, and the UGV uses a VLM to augment semantic descriptions of the environment.

\section{Lessons Learned}

\subsection{Importance of Robust Communication}

One of the restrictions of using server-based LLMs for planning is that the robot requires continuous communication with the server. As described in \S\ref{subsec:comms}, we rely on mesh radios to ensure that robots can execute their API calls on the field. It may be challenging to ensure internet connectivity during field experiments, requiring the use of cellular connection or satellite internet.
Deploying and testing such mesh networks requires setup time, carrying relay nodes to strategic locations, and configuring different non-overlapping communication frequencies between robots. In these scenarios, \emph{Mobile Infrastructure on Demand}~\cite{mox2024opportunistic} may provide solutions that ensure network connectivity between the different robots.
Inevitably, robots may eventually leave the communication range. In this case, it is advantageous that the autonomy stack enables operations without communications, enabling the robot to return to an area within communication range after a period of isolation.

\subsection{Modular Autonomy}

\glspl{llm} have enabled roboticists to move beyond structured tasks in controlled settings by providing common sense knowledge about the world.
This allows robots to reason more flexibly and handle dynamic scenarios.
As the field advances, multimodal models, which integrate sensory inputs (\emph{e.g}., vision, touch, language), offer even greater potential by directly generating plans from rich perceptual data.
However, while this direct mapping is often desirable, it carries risks.
These models can occasionally produce infeasible or unsafe plans with little transparency into their reasoning.
Our results (Fig.~\ref{fig:validation}) demonstrate the value of interpretability.
Even minimal feedback—such as explaining why a plan was infeasible (e.g., disconnected nodes in the planning graph)—significantly improved performance.
This highlights the need for validation mechanisms to enhance reliability.
To mitigate the risks of opaque, end-to-end LLM-based planning, we advocate for a modular autonomy framework.
Retaining access to well-understood, low-level behaviors (\emph{e.g}., obstacle avoidance, motion primitives) allows for closed-loop refinement.
By coupling traditional autonomy algorithms with LLM-based decision-making, the system can balance broad reasoning capabilities with safe, interpretable execution.
Without this hybrid approach, \glspl{llm} and traditional autonomy may struggle to work together effectively, limiting the system’s reliability in real-world tasks.

\section{Open Challenges}

In the following section we discuss open challenges in deploying FM-enabled robots in the field.

\subsection{LLMs on-the-edge}
While most LLM-enabled robots rely on GPT-4 or a similar LLM that requires offboard cloud-compute, robots in the field are often network-denied.
Technologies such as satellite internet provide one approach for localized internet access,
however this solution requires additional infrastructure that may limit the range of robots in the field.
Deploying LMs directly onboard is a promising approach towards enabling robust contextually-aware robots in the field, but doing so is often not straightforward.
A modern onboard GPU may have anywhere from 5 GB to 20 GB of VRAM, and this limits the class of models that may be used.
Due to the capability gap between classes of LMs, the LM-enabled planner will likely suffer significant performance degradation.
Model distillation is one promising path towards bridging the capability gap between classes of LLMs, as we demonstrate above.
In the context of robot planning, there are several possible distillation directions, which we discuss below.

\medskip

\noindent \textbf{LMs for full task planning.} The most direct approach to LM distillation is to train a smaller model to directly mimic the behavior of an ``expert'' larger model, which entails training a planning model such as those used in SPINE~\cite{ravichandran_spine}.
This approach has the advantage of producing a drop-in replacement, which requires no other changes in the planning architecture.
However, there are cases where even this approach may be prohibitively resource intensive.

\medskip

\noindent \textbf{LM for partial task planning.} In cases where distilling a fully functional LM may not be feasible  (\textit{e.g.}, due to compute limitations), a second option may be to leverage approaches to task decomposition.
In this paradigm, a more powerful LM (\textit{e.g.}, GPT-4o) decomposes a complex mission specification into smaller, more manageable tasks.
Smaller LMs then execute the smaller tasks.
In this regime, heterogeneous teams could be allocated tasks that are not just limited by their physical capabilities but also their compute capabilities.
This is further exacerbated when autonomy algorithms with strict latency requirements must also be run onboard.
In our example, the \glspl{uav} may only handle tasks that are reasonably well specified such as ``map a parking lot" rather than require the ability to handle more complex tasks that require implied reasoning such as ``where am I likely to find a car".
This may alleviate the burden of training data and the required size of the models that need to be run onboard.

\subsection{Visual Foundation models on-the-edge}

Visual foundation models, including \glspl{vlm} and grounded detection models, provide powerful open-vocabulary scene understanding.
However, because VLMs are primarily trained on data from indoor and structured environments, using them in the field incurs a distribution shift that causes performance degradation, particularly in non-traditional viewpoints such as that of an aerial vehicle.

\begin{figure}
    \centering
    \includegraphics[width=0.9\linewidth]{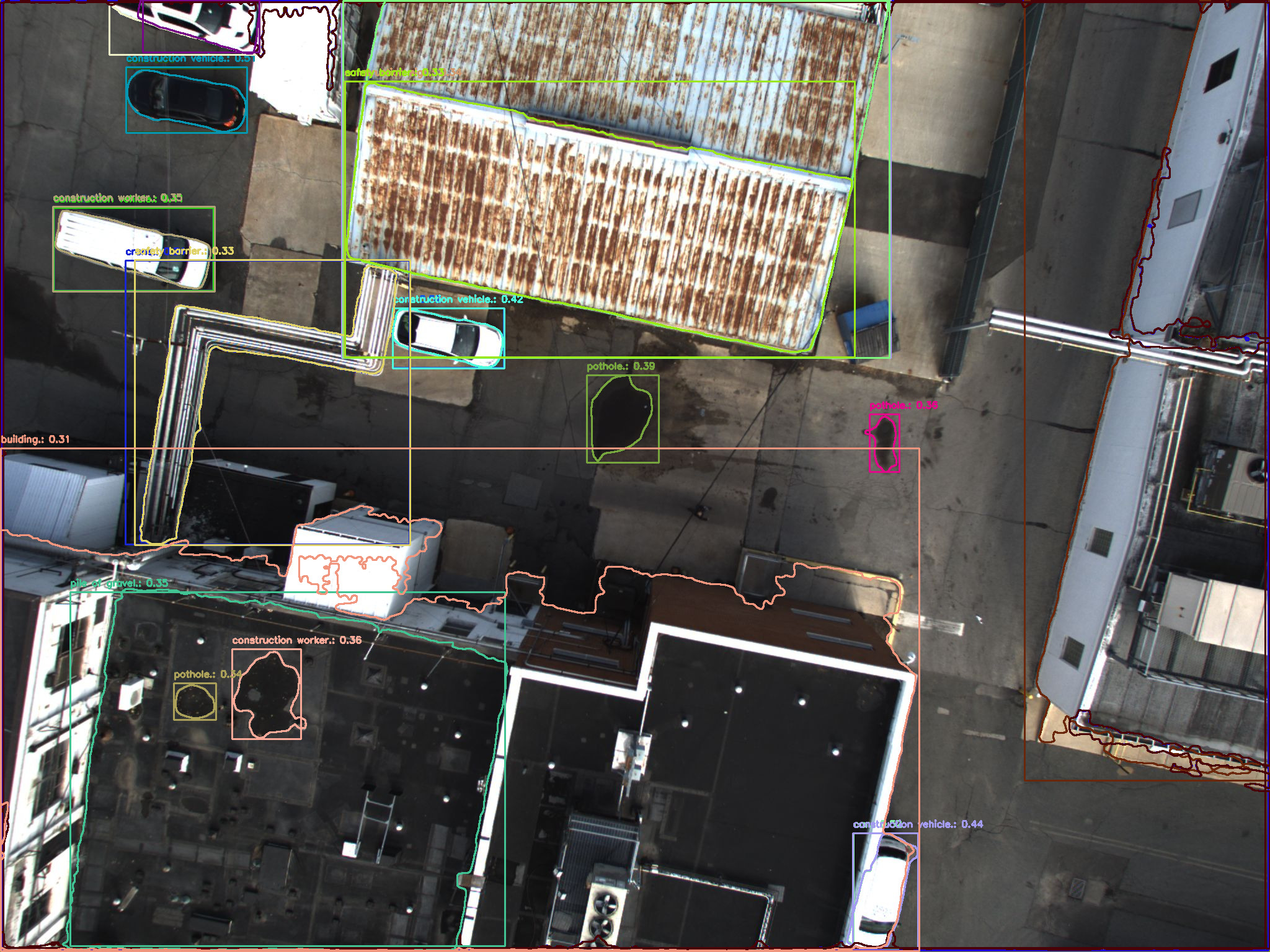}
    \caption{Example of the shortcomings of a Grounded Detection model. Cars a labeled as construction vehicles because the task is relevant to finding road work. Figure from~\cite{cladera2025tfr}.}
    \label{fig:grounded-dino}
    \vspace{-.3cm}
\end{figure}

As discussed in \S\ref{sec:air-ground-teaming}, we use a grounded segmentation model to construct graphs of objects and traversable areas. Consider a scenario where we ask the system to inspect roads for construction. Some classes it looks for are \texttt{construction vehicle}, \texttt{pothole}, and \texttt{safety barrier}. As shown in Fig. \ref{fig:grounded-dino}, the grounded model detects regular cars as construction vehicles and buildings as safety barriers. We see that the
model cannot distinguish between specialized objects for the task and normal objects, which we would expect the UAV to see as it navigates the environment. In the context of a single image, one can see how the model may be more confident that a car is a construction vehicle, especially if car is not an alternative class, but in the context of the whole environment, it should not be.
Additionally, these models suffer from false positives, such as the building labeled as a safety barrier and a puddle labeled as a construction worker. These shortcomings could be mitigated through fine-tuning, but this could inhibit generalization.

In addition to grounded detection models, VLMs also have shortcomings, particularly related to model size. In order to deploy these models on robots we tend to use smaller models with $\leq$7B parameters. Consider a scenario where we had the robot look for injured people. When using a 70B parameter model, the model would describe seeing blood on the person's clothes and say that the person was injured. When testing a 7B parameter model, it would describe seeing blood and then say the person was not injured.
Our empirical observation is that smaller models tend to lack the comprehension that larger models have.

\subsection{Evaluation \& Datasets for Contextual Planning at Scale}

One of the major selling points of LLM-enabled robots is their ability to execute complex mission specifications expressed in natural-language.
While the community is developing evaluation procedures for language-driven planning and control in other domains (\textit{e.g.}, manipulation~\cite{li24simpler}), to the best of our knowledge, there are no evaluation protocols for language-specified missions that require large-scale navigation and mapping in unstructured environments.
While large datasets with language exist~\cite{open_x_embodiment_rt_x_2023} and internet data has been used to translate ego-centric videos to robotic applications~\cite{sanpo_dataset}, there is still a large gap in standardized evaluation procedures for outdoor unstructured robot navigation in \emph{open-world} and \emph{open-task} settings.
The community has made progress in indoor scenes~\cite{xia2018gibson, yokoyama2024hm3d, chang2017matterport3d, deitke2020robothor} with both well specified tasks~\cite{ramakrishnan2021habitat} and conversational tasks that may require reasoning~\cite{he2021landmark}.
For empirical rigor, we need to find widely accepted missions that describe both the set of tasks and the environment descriptions that robots may find themselves in.
This would allow us to establish standard evaluation protocols for benchmarking these algorithms.
We briefly outline the desiderata for such an evaluation procedure.

\medskip

\noindent \textbf{Mission Complexity.} This procedure should contain mission specifications of increasing complexity across dimensions such as the number of steps required to complex a mission, the difficulty of the semantic associations required, or the amount of exploration required to complete a mission.

\medskip

\noindent \textbf{Environmental Diversity.} One of the key challenges of field robotics is the diverse and unpredictable nature of environments a robot must face.
The field robotics community has successfully deployed robots in forests, mountains, snowy regions, and tunnels ~\cite{liu2022large, lee2020learning, LaRocque2024, khattak2020nebula}. The evaluation must reflect this diversity both visually and semantically.

\medskip

\noindent \textbf{Platforms.} Field roboticists employ  wheeled vehicles, quadrupeds, UAVs, UUVs, and other platforms.
Effective evaluation procedures would be applicable to all of these platforms.
LLM-enabled planners typically consider high-level control --- abstracting aware dynamics in favor of waypoint targets or similar --- which naturally accommodates various platforms.
This problem is more difficult for lower-level control approaches, such as those proposed by VLAs.

\section{Conclusion}

While FM-enabled robots offer a natural human interface and advanced contextual reasoning abilities,  existing FM-enabled planning approaches provide limited utility for field robotic applications.
This paper discusses our recent efforts to overcome these limitations through  our autonomy framework powered by SPINE, an LLM-enabled planner capable of operating in partially-known and unstructured environments.
We use this framework to accomplish large-scale missions in unstructured environments.
We use our framework to distill lightweight language models, which we use for onboard UAV planning.
Finally, we discuss lessons learned during our deployments and outline promising directions for future research.

\bibliographystyle{IEEEtran}
\bibliography{IEEEabrv, literature, zac_refs}

\end{document}